\documentclass[sigconf]{acmart}

\usepackage{algorithm}
\usepackage{algorithmic}
\usepackage{amsmath}
\usepackage{booktabs}
\usepackage{color}
\usepackage{graphics}
\usepackage{multirow}

\def \h {\mathbf{h}}
\def \u {\mathbf{u}}

\def \p {\mathbf{p}}
\def \r {\mathbf{r}}
\def \T {\mathbf{T}}
\def \Y {\mathbf{Y}}
\def \R {\mathbb{R}}
\def \W {\mathbf{W}}

\def \M {\mathcal{M}}
\def \P {\mathcal{P}}
\def \CD {\mathcal{D}}
\def \CI {\mathcal{I}}
\def \CL {\mathcal{L}}
\def \CS {\mathcal{S}}

\def \CLS {\mbox{CLS}}
\def \SEP {\mbox{SEP}}
\def \Action {\texttt{Action}}
\def \Argument {\texttt{Argument}}
\def \Problem {\texttt{Problem}}
\def \Question {\texttt{Question}}

\def \FUNCTION {\textbf{Function}}
\def \ENDFUNCTION {\textbf{EndFunction}}
\def \RET {\textbf{return}}

\newcommand{\cparagraph}[1]{\noindent\textbf{#1}}

\AtBeginDocument{%
  \providecommand\BibTeX{{%
    \normalfont B\kern-0.5em{\scshape i\kern-0.25em b}\kern-0.8em\TeX}}}
\setcopyright{iw3c2w3}
\copyrightyear{2021}
\acmYear{2021}
\acmDOI{10.1145/3442381.3450026}

\acmConference[WWW '21]{Proceedings of the Web Conference 2021}{April 19--23, 2021}{Ljubljana, Slovenia}
\acmBooktitle{Proceedings of the Web Conference 2021 (WWW '21), April 19--23, 2021,
Ljubljana, Slovenia}
\acmPrice{}
\acmISBN{978-1-4503-8312-7/21/04}
\begin{document}

%%
%% The ''title'' command has an optional parameter,
%% allowing the author to define a ''short title'' to be used in page headers.
\title{Automatic Intent-Slot Induction for Dialogue Systems}

%%
%% The ''author'' command and its associated commands are used to define
%% ''authornote'' and ''authornotemark'' commands
\author{Zengfeng Zeng}
\email{zengzengfeng277@pingan.com.cn}
\affiliation{%
  \institution{Ping An Life Insurance of China, Ltd.}
  %\streetaddress{5033 Yitian Rd}
  %\city{Shenzhen Shi}
  %\state{Guangdong Sheng}
  \country{China}}
  
\author{Dan Ma}
\authornote{Corresponding author.}
\email{whmadan1990@gmail.com}
\affiliation{%
  \institution{Ping An Life Insurance of China, Ltd.}
  %\streetaddress{5033 Yitian Rd}
  %\city{Shenzhen Shi}
  %\state{Guangdong Sheng}
  \country{China}}
  
\author{Haiqin Yang}
\email{hqyang@ieee.org}
\affiliation{%
  \institution{Ping An Life Insurance of China, Ltd.}
  %\streetaddress{5033 Yitian Rd}
  %\city{Shenzhen Shi}
  %\state{Guangdong Sheng}
  \country{China}}
  
\author{Zhen Gou}
\email{gouzhen508@pingan.com.cn}
\affiliation{%
  \institution{Ping An Life Insurance of China, Ltd.}
  %\streetaddress{5033 Yitian Rd}
  %\city{Shenzhen Shi}
  %\state{Guangdong Sheng}
  \country{China}}
  
\author{Jianping Shen}
\email{shenjianping324@pingan.com.cn}
\affiliation{%
  \institution{Ping An Life Insurance of China, Ltd.}
  %\streetaddress{5033 Yitian Rd}
  %\city{Shenzhen Shi}
  %\state{Guangdong Sheng}
  \country{China}}

\renewcommand{\shortauthors}{Zengfeng Zeng, Dan Ma, Haiqin Yang, Zhen Gou and Jianping Shen}

%% article.labeling coarse-grained intent roles to quadruples via sequence labeling (assigning each intent-role mention to fine-grained concepts that are mined from open-domain corpus via clustering ( abstracting fine-grained concepts for the extracted intent-role mentions via clustering
\begin{abstract}
Automatically and accurately identifying user intents and filling the associated slots from their spoken language are critical to the success of dialogue systems.  Traditional methods require manually defining the DOMAIN-INTENT-SLOT schema and asking many domain experts to annotate the corresponding utterances, upon which neural models are trained.  This procedure brings the challenges of information sharing hindering, out-of-schema, or data sparsity in open domain dialogue systems.  To tackle these challenges, we explore a new task of {\em automatic intent-slot induction} and propose a novel domain-independent tool.  That is, we design a coarse-to-fine three-step procedure including Role-labeling, Concept-mining, And Pattern-mining (RCAP): (1) role-labeling: extracting key phrases from users' utterances and classifying them into a quadruple of coarsely-defined intent-roles via sequence labeling; (2) concept-mining: clustering the extracted intent-role mentions and naming them into abstract fine-grained concepts; (3) pattern-mining: applying the Apriori algorithm to mine intent-role patterns and automatically inferring the intent-slot using these coarse-grained intent-role labels and fine-grained concepts.  Empirical evaluations on both real-world in-domain and out-of-domain datasets show that: {(1) our RCAP can generate satisfactory SLU schema and 
outperforms the state-of-the-art supervised learning method; (2) our RCAP can be directly applied to out-of-domain datasets and gain at least 76\% improvement of F1-score on intent detection and 41\% improvement of F1-score on slot filling; (3) our RCAP exhibits its power in generic intent-slot extractions with less manual effort, which opens pathways for schema induction on new domains and unseen intent-slot discovery for generalizable dialogue systems. } 

\end{abstract}




\maketitle
\begin{figure}
  \includegraphics[scale=0.40]{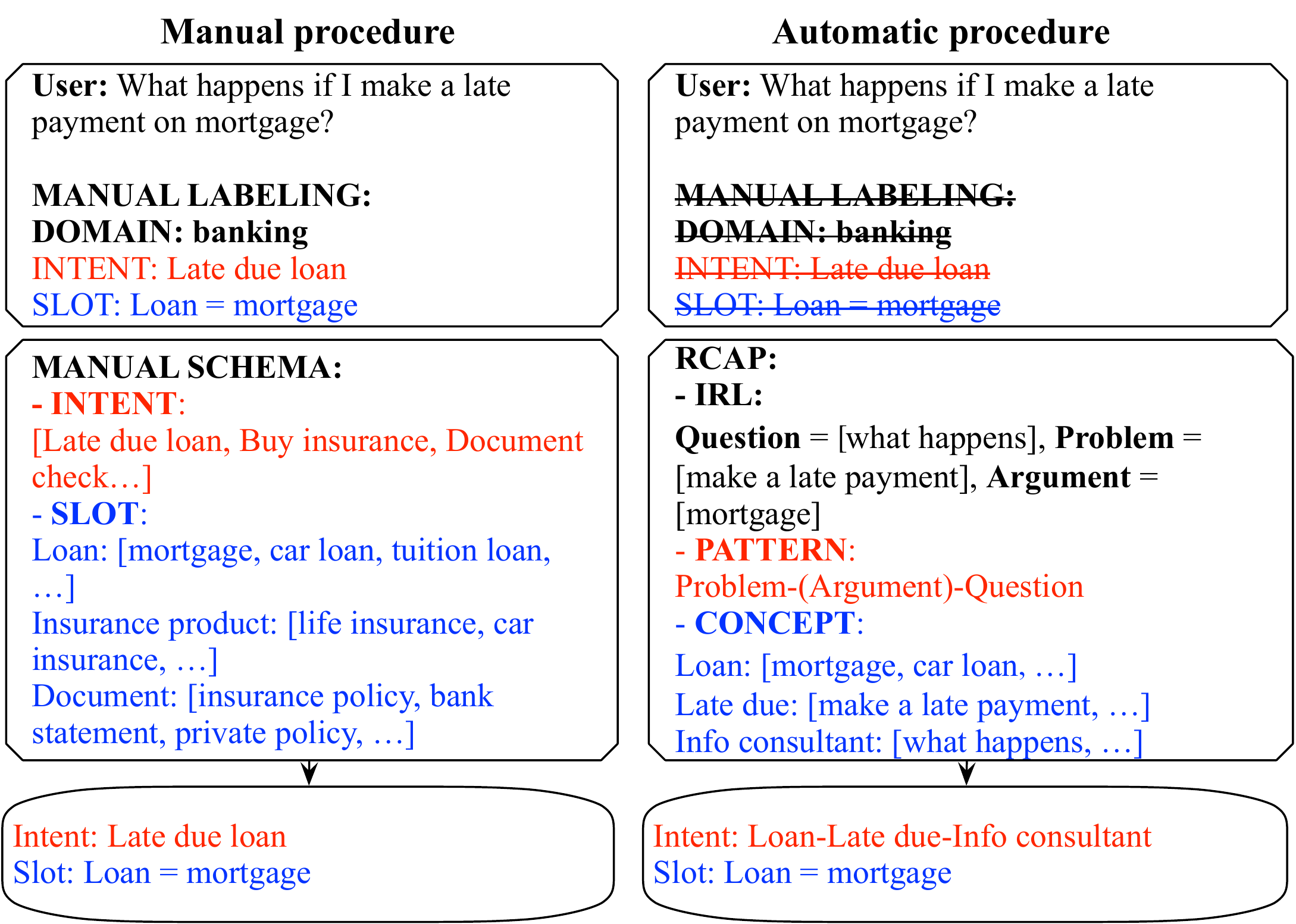}
  \centering
  \caption{
  Comparison between traditional manual intent-slot construction and automatic induction. The traditional procedure requires domain experts to manually  annotate utterances into the DOMAIN-INTENT-SLOT schema (see the left-top box) and many manually annotated schemas (see the left-middle box) while our RCAP can automatically infer the intent-slot without manual labeling.
  }  \label{fig:running_example} 
\end{figure} 

\section{Introduction}
{Recently, thanks to the advance of artificial intelligence technologies and abundant real-world conversational data, virtual personal assistants (VPAs), such as Apple's Siri, Microsoft's Cortana and Amazon's Alexa, have been developed to help people's daily life~\cite{chen2018deep}. Many VPAs have incorporated task-oriented dialogue systems to emulate human interaction to perform particular tasks, e.g., customer services and technical supports~\cite{chen2017survey}.}  

Spoken language understanding (SLU) is a crucial component to the success of task-oriented dialogue systems~\cite{DBLP:conf/naacl/DamonteGC19,tur2011spoken}.  Typically, a commercial SLU system detects the intents of users' utterances mainly in three steps~\cite{goo-etal-2018-slot,kim2016intent,coucke2018snips,chen-etal-2013-identifying}: (1) identifying the dialogue domain, i.e., the area related to users' requests; (2) predicting users' intent; and (3) tagging each word in the utterance to the intent’s slots.  To appropriately solve this task, traditional SLU systems need to learn a model from a large predefined DOMAIN-INTENT-SLOT schema annotated by domain experts and professional annotators.  For example, as illustrated in Fig.~\ref{fig:running_example}, given a user's utterance, ``What happens if I make a late payment on mortgage?'', we need to label the domain to ``banking'', the intent to ``Late due loan'', and the slot ``Loan'' to ``mortgage''.  

The annotation procedure usually requires many domain experts to conduct the following two steps~\cite{chen2013unsupervised,wang2012crowdsourcing}: (1) selecting related utterances from specific domains based on their domain knowledge; (2) examining each utterance and enumerating all intents and slots in it.  This procedure, however, faces several critical challenges.  First, it is redundant as experts cannot effectively share the common information among different domains.  For example, given two utterances ``Can I check my insurance policy?'' and ``Can I read my bank statement?'' from the domains of  banking and insurance, their intents can be abstracted into ``check document''.  They are usually annotated by at least two experts from different domains, which hinders the information sharing.  Second, the labeling procedure may be biased to experts' knowledge and limited by their domain experience.  To meet more users' needs, dialogue systems usually have to cover a number of domains and solicit sufficient domain experts to build comprehensive schemas.  The requirement of domain experts increases the barrier of scaling up the dialogue systems.  Third, it is extremely hard to enumerate all intents and slots in the manual procedure.  Usually, the intent-slot schema follows the long-tail distribution.  That is, some intents and slots rarely appear in the utterances.  Experts tend to ignore part of them due to the human nature of memory burden.  Fourth, for system maintenance, it is nontrivial to determine whether there are new intents or not in a given  utterance.  Hence, experts have to meticulously examine each utterance to determine whether new intents and slots exist.

To tackle these challenges, researchers have incorporated different mechanisms, such as crowd sourcing~\cite{wen2017network} and semi-supervised learning~\cite{touretzky1996advances}, to assist the manual schema induction procedure.  They still suffer from huge human effort.  Other work further applies unsupervised learning techniques to relieve the manual effort~\cite{chen2014leveraging,DBLP:conf/acl/LiuWXF20,vedula2020open,shi-etal-2018-auto}.  For example, unsupervised semantic slot induction and filling~\cite{chen2014leveraging,DBLP:conf/acl/LiuWXF20} have been proposed accordingly.  However, they cannot derive intents simultaneously.  Open intent extraction has been explored~\cite{vedula2020open} by restricting the extracted intents to the form of \emph{predicate-object}.  It does not extract slots simultaneously.  Moreover, a dynamic hierarchical clustering method~\cite{shi-etal-2018-auto} has been employed for inducing both intent and slot, but can only work in one domain.  
   
\begin{figure*}
  \includegraphics[scale=0.35]{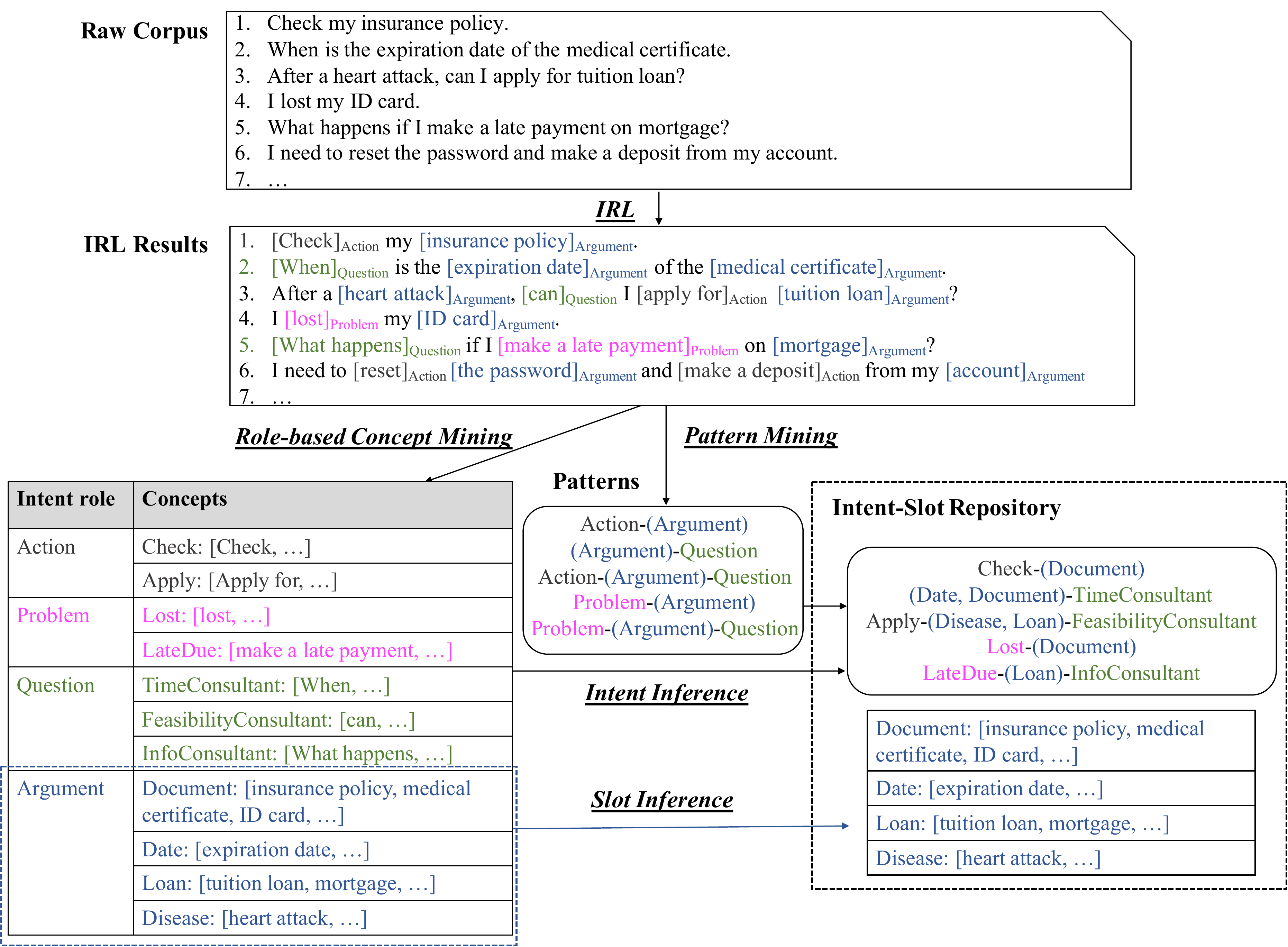}
  \centering
  \caption{Flow of RCAP.  The intent-role mentions and concepts are highlighted by different colors: $\Argument$ in blue, $\Action$ in gray, $\Problem$ in magenta, and $\Question$ in green.  Mined concepts on each intent-role are shown in square brackets in the left-bottom table.  The mined intent-role patterns are order-irrelevant.  A round-bracket in $\Argument$ implies no mention or several mentions.  
  }  
  \label{fig:patterns}
\end{figure*}

In this paper, we define and investigate a new task of automatic intent-slot induction (AISI).  We then propose a coarse-to-fine three-step procedure, which consists of Role-labeling, Concept-mining, And Pattern-mining (RCAP).  The first step of role-labeling comes from the observation of typical task-oriented dialogue systems~\cite{tesniere1959elements,white2015questions,kollar2018alexa,liu2020sentiment,di2015development} that 
utterances can be decomposed into a quadruple of coarsely-defined intent-roles: $\Action$, $\Argument$, $\Problem$, and $\Question$, which are independent to concrete domains.  Thus, we build an intent-role labeling (IRL) model to automatically extract corresponding intent-roles from each utterance.  By such setting, as shown in Fig.~\ref{fig:patterns}, we can determine the utterance of ``Check my insurance policy'' to $\Action=[\mbox{Check}]$ and $\Argument=[\mbox{insurance policy}]$ while the utterance of ``I lost my ID card'' to $\Problem=[\mbox{lost}]$ and $\Argument=[\mbox{ID card}]$.  Secondly, to unify utterances within the same intent into the same label, as shown in  Fig.~\ref{fig:tsne_results}.  We deliver concept mining by grouping the mentions within the same intent-role and assigning each group to a fine-grained concept.  For example, the mentions of ``insurance policy'', ``medical certificate'', and ``ID card'' in $\Argument$ can be automatically grouped into the concept of ``Document'' while the mentions of ``tuition loan'' and ``mortgage'' can be grouped into the concept of ``Loan''.  Here, we only consider one-intent in one utterance, which is a typical setting of intent detection in dialogue systems~\cite{liu2016attention}.  Hence, multi-intent utterances, e.g., ``I need to reset the password and make a deposit from my account.'', are excluded.  Thirdly, to provide intent-role-based guidelines for intent reconstruction, we conduct Apriori~\cite{yabing2013research} and derive the intent-role patterns, e.g., the Patterns in Fig.~\ref{fig:patterns}.  Specifically, the extracted intent-roles are fed into Apriori to obtain frequent intent-role patterns, e.g., $\Action$-$(\Argument)$.  Finally, we combine the mined concepts according to the intent-role patterns to derive the intent-slot repository.  For example, as illustrated in Fig.~\ref{fig:patterns}, given an utterance of ``Check my insurance policy'', according to the obtained pattern of $\Action$-$(\Argument)$, we can assign the concepts to it and infer the intent of ``Check-(Document)'' with ``insurance policy'' in the slot of ``Document''.

In the literature, there is no public dataset to be applied to verify the performance of our proposed RCAP.  Though existing labeled datasets, such as ATIS~\cite{price1990evaluation} and SNIPS~\cite{coucke2018snips}, have provided concise, coherent and single-sentence texts for intent detection, they are not representative for complex real-world dialogue scenarios as spoken utterances may be verbose and ungrammatical with noise and variance~\cite{tur2010left}.  Hence, we collect and release a financial dataset (FinD), which consists of 2.9 million real-world Chinese utterances from nine different domains, such as insurance and financial management.  Moreover, we apply RCAP learned from FinD to two new curated datasets, a public dataset in E-commerce and a human-resource dataset from a VPA, to justify the generalization of our RCAP in handling out-of-domain data.

We summarize the contributions of our work as follows: 
\begin{itemize}
\item[--]  We define and investigate a new task in open-domain dialogue systems, i.e., automatic intent-slot induction, and propose a domain-independent tool, RCAP.  
\item[--]  Our RCAP can identify both coarse-grained intent-roles and abstract fine-grained concepts to automatically derive the intent-slot.  The procedure can be efficiently delivered.
\item[--]  More importantly, RCAP can effectively tackle the AISI task in new domains.  This sheds light on the development of generalizable dialogue systems.
\item[--]  We curate large-scale intent-slot annotated datasets on financial, e-commerce, and human resource and conduct experiments on the datasets to show the effectiveness of our RCAP in both in-domain and out-of-domain SLU tasks.
\end{itemize}

\section{Problem Formulation}
\label{sec:task_formulation}
The task of automatic intent-slot induction is defined as follows: given a set of raw utterances, $\CD=\{\u_i\}_{i=1}^M$, where $M$ is the number of utterances in the set, $\u_i=u_{i_1}\ldots u_{i_{|u_i|}}$ denotes an utterance with $|u_i|$ sub-words, our goal is to derive its intent $\CI_{{i}}$ for the corresponding utterance $\u_i$.  Here, we only consider one intent in one utterance, which is a typical setting of intent detection in dialogue systems~\cite{liu2016attention}.  Since each intent has its corresponding slots, we set the slots of $\CI_i$ as $\CS_i=\{\CS_{i,1}, \ldots, \CS_{i,\CL_i}\}$, where $\CL_{i}$ is the number of slots in $\CI_i$ and $\CS_{i,j}$ is a tuple with the name and value, $j=1,\ldots, \CL_{i}$.  It is noted that $\CL_{i}$ can be 0, implying no slot in the intent.  For example, ``How to get insured?'' contains the intent of ``Buy insurance'' without a slot.

In our work, the intents are dynamically decided by the procedure of Intent Role Labeling, Concept Mining, and Intent-role Patterns Mining.  $\CS_i$ is also learned automatically and dynamically.  
 
To provide a domain-independent expression of intents, we follow~\cite{liu2020sentiment} and decompose an utterance into several key phrases with the corresponding intent-roles defined as follows:
\begin{definition}\label{def:irl}%[Intent-role \red{labels}]
An intent-role is a label from the following set:
\begin{equation}\label{eq:irl_set}
\left\{{\Action}, {\Argument}, {\Problem}, {\Question}\right\},
\end{equation}
where ${\Action}$ is a verb or a verb phrase, which defines an action that the user plans to take or has taken.  ${\Question}$ delivers interrogative words or an interrogative phrase, which defines a user's intent to elicit information.  ${\Problem}$ outlines a failure or a situation that does not meet a user's expectation.  ${\Argument}$ expresses in nouns or noun phrases to describe the target or the holder of ${\Action}$ or ${\Problem}$.
\end{definition} 

To further provide fine-grained semantic information for each intent-role mention, we define concepts as~\cite{chen2013unsupervised}:
{
\begin{definition}[Concept]\label{def:concept}
Given the extracted intent-role mentions, we can individually and independently group the mentions within each intent-role and name each cluster by a concept, an abstraction of similar instances. 
  
\end{definition}
}

To rationally combine the concepts under each intent-role and reform the user intent, we define intent-role pattern as follows:

\begin{definition}[Intent-role Pattern]
For each utterance, we decompose it into several intent-role mentions.  A combination of intent-roles is defined by an intent-role pattern. 
\end{definition}

In this paper, we propose RCAP to tackle the task of AISI.  Our RCAP consists of three modules: (1) intent-role labeling for recognizing the intent-roles of mentions, (2) concept mining for fine-grained concepts assignment, and (3) intent-role pattern mining to attain representative patterns.  After that, we can infer the intent-slot accordingly.

\section{Our Proposal}
In this section, we present the implementation of the modules of our RCAP.  

\subsection{Intent Role Labeling (IRL)}\label{sec:IRL}
In order to attain coarse-grained intent-roles as defined in Def.~\ref{def:irl}, we train an IRL model on an open-domain corpus with $L$ annotated utterances.  That is, given an utterance with $m$ sub-words, $\u = u_1\ldots u_{m}$, we train an IRL model to output the corresponding label, $\r=r_1\ldots r_{m}$.  Here, we apply the Beginning-Inside-Outside (BIO) schema~\cite{ramshaw1999text} on the four intent-roles.  Hence, $r_i$ can be selected from one of the 9 tags, such as B-$\Action$ and I-$\Argument$.

Nowadays, BERT~\cite{kenton2019bert} has demonstrated its superior performance on many downstream NLP tasks.  We, therefore, apply it to tackle the IRL task. More specifically, given the utterance $\u$, we can denote it by 
\[
[\CLS]\,u_1\,\ldots\,u_{m}\,[\SEP],
\]
where $[\CLS]$ and $[\SEP]$ are two special tokens for the classification embedding and the separation embedding.  $u_i$ is the $i$-th subword extracted from BERT's dictionary.  By applying BERT's representation, we obtain
\[
\h_0\,\h_1\,\ldots\,\h_{m}\,\h_{m+1},
\]
where $\h_i\in\R^d$ is the embedding of the $i$-th token, $i=0,\ldots, m+1$ and $d$ is the hidden size.  We then apply a softmax classifier on top of the hidden features to compute the probability of each sub-word $u_{i}$ to the corresponding label: %$r_{j}$  ($j=1, 2, 3, 4$):
\begin{equation}\label{eq:irl}
p({r}_{i}|u_{i}) = \mbox{softmax}(\W \h_{i}),
\end{equation}
where $\W$ is the weight matrix.  After that, mentions are obtained by the BI-tags on each intent-role.  

\subsection{Concept Mining}
\label{sec:concept_mining}
The goal of concept mining is to provide fine-grained labels for the determined intent-role mentions obtained in Sec.~\ref{sec:IRL}.  To attain such goal, we group the mentions within the same intent-role into clusters and assign each cluster to the corresponding \emph{concept} (see Def.~\ref{def:concept}) by a fine-grained label.  There are two main steps: mention embeddings and mention clustering.  After that, we can assign abstract fine-grained names for the clusters.

\cparagraph{Mention Embedding} This step takes the sub-word sequence of all mentions in the open domain utterances $\CD^r_{m}=\{m_{kr}\}_{k=1}^{M_r}$ and outputs the embedding vector $\p_{kr}$ for each mention $m_{kr}$, where $M_r$ is the number of mentions in the corresponding role, $r\in\{\Action, \Question, \Argument, \Problem\}$.  There are various ways to represent the intent-role mentions.  To guarantee unified representations of all mentions, we do not apply BERT because its representation will change with the context. Differently, we consider the following embeddings:
\begin{itemize}
\item[--]  {\bf word2vec (w2v):} It is a popular and effective embedding in capturing semantic meanings of sub-words.  We treat intent-role mentions as integrated sub-words and represent them following the same procedure in~\cite{mikolov2013efficient}.
\item[--]  {\bf phrase2vec (p2v):} To further include contextual features, we not only take intent-role mentions as integrated sub-words but also apply phrase2vec~\cite{artetxe2018unsupervised}, i.e., a generalization of skip-gram to learns n-gram embeddings. 
\item[--]  {\bf CNN embedding (CNN):} To make up the insufficiency of word2-vec and phrase2vec in sacrificing semantic information inside mentions, we apply a sub-word convolutional neural network (CNN)~\cite{zhang2015character} to learn better representations.  That is, a CNN model takes the sequence of an input mention and outputs an embedding vector $\p$ by applying max pooling along the mention size on top of the consecutive convolution layers.  The CNN embedding model is trained by skip-gram in an unsupervised manner~\cite{mikolov2013efficient}.  Given mention $t$ extracted from an utterance in the training set, we seek the embedding by minimizing the following loss for each mention $c$ within the context of $t$:
\begin{equation}
L_p = -[\log\sigma (\p_c^{T}\p_t) + \sum^{M}_{i=1}{\log\sigma(-\p_i^{T}\p_t)}],
\end{equation}
where $\sigma$ is the sigmoid function,  $\p_t$ and $\p_c$ are embeddings for the mention $t$ and $c$, respectively. $\p_i$ is the embedding of a random mention that does not occur in the context of mention $t$ and $M$ is the number of randomly selected mentions. 
\end{itemize}

\cparagraph{Mentions Clustering} 
After obtaining the mention embeddings, we apply clustering on the mentions within the same intent-role to group them into corresponding concepts.  In this paper, we apply the following algorithms: 

\begin{itemize}
\item[--]  {\bf K-means}~\cite{kanungo2002efficient}: It is one of the most popular clustering algorithms.  However K-means algorithm needs to decide initial centroids and preset the number of clusters in advance. 
\item[--]  {\bf Minimum entropy (MinE)}~\cite{rosvall2008maps}: It is a famous algorithm by Minimizing Entropy on infomap.  They apply a transition probability matrix to discover connected structure and have proved the effectiveness in community detection.  %Here, we use this algorithm for better clustering performance.
\item[--]  {\bf Label Propagation Algorithm (LPA)}~\cite{raghavan2007near}: It is an effective clustering algorithm and does not need to specify the number of clusters in advance.  Here, we construct an $n\times n$ transition matrix $\T$ through the learned phrase embeddings, where $n$ is the number of phrases and $T_{ij}=\p_i^T\p_j$ defines the inner product of two row-wise normalized vectors, $\p_i$ and $\p_j$.  To take into account the $k$-nearest neighbors in vector space, we only keep the top-$k$ elements in each row of $\T$.

Initially, we consider each phrase as a cluster and initialize $\Y$ by an identity matrix, where $Y_{ij}$ denotes the probability that phrase $i$ belongs to the cluster $j$.  We then update $\Y$ by:
\begin{equation}
\Y_{t} = \T\Y_{t-1}.
\end{equation}
After convergence, each phrase is assigned to the cluster with the highest probability.  
%Several sample concepts are included in appendix Fig. 4.
\end{itemize}

\subsection{Intent-role Pattern Mining}
\label{sec:pattern_mining}
To reconstruct intents from extracted intent-role mentions and concepts, we aim to explore the common patterns that people express their intents.  Each pattern is a combination of intent-roles without considering the order.  It is noted that by enumerating all combinations of intent-roles in Eq.~(\ref{eq:irl_set}), we can obtain 15 candidate patterns, which is computed by $\tbinom{4}{1}+\tbinom{4}{2}+\tbinom{4}{3}+\tbinom{4}{4}=4+6+4+1=15$. 

Given a large corpus with intent-roles as defined in Def.~\ref{def:irl}, we then apply Apriori algorithm~\cite{yabing2013research}, a popular frequent item set mining algorithm, to extract the most frequent intent-role combination patterns. The corresponding parameters, such as the minimum support value and the minimum confidence value can be adjusted. 

Hence, given an utterance, we apply the learned IRL model to identify the mentions with intent-roles.  According to Sec.~\ref{sec:intent_slot_induction}, we can map the mentions to appropriate concepts and determine the corresponding intent-slot based on the mined intent-role patterns.  

\section{Inference in SLU} 
\label{sec:intent_slot_induction}
In the following, we outline how to apply the learned IRL model, concepts, and intent-role patterns by our RCAP for real-world AISI tasks.  Algorithm~1 outlines the procedure of inferring the intent-slot by our RCAP:
\begin{itemize}
\item[--]  Line~1 is to apply the learned IRL model, $\mbox{\bf IRL}$, to extract the mentions with the corresponding intent-roles for the given utterance $\u$.  For example, given an utterance, ``Check my medical report'', we obtain two meaningful mention-role pairs, ({``Check''}, $\Action$) and ({``medical report''}, $\Argument$).  
\item[--]  Line~2 is to invoke $ConInfer$ to assign each mention to a suitable concept within the same intent-role.  Here, we first assign the concept by direct matching.  $ConInfer$ in line~6-20 lists the procedure when no exact mention appears in the concept set of $\M$.  Specifically, we compute the cosine similarity between the mention and the mentions of all concepts within the same intent-role.  We then attain the concept IDs of the top-$K$ neighbors of the mention and apply the majority vote to determine the concept.  Since $m=$``medical report'' does not appear in the mention of $\M$, we first compute the cosine similarity between $m$ and all mentions within $\Argument$ in $\M$.  By finding the majority concept from the top-$K$ neighbor mentions, we assign ``Document'' to the concept of ``medical report''.  If no concept is matched, we run the procedure of {\bf Concept Expansion} as in Appendix~\ref{sec:SI-concept-exp}.
\item[--]  Line~3 is to invoke $ISInfer$ to derive the final intent-slot as defined in line~22-36.  It is noted that line~25-27 is to extract multiple slots.  For example, the utterance, ``When is the expiration date of the medical certificate'', contains multiple slots including  ``expiration date'' and  ``medical certificate''.  An intent is then obtained by concatenating all intent-roles filled with corresponding concepts as in line 33.  Hence, by filling the concept of ``Check'' to $\Action$ and the concept of ``Document'' to $\Argument$, we obtain the intent of ``Check-(Document)'' for ``Check my medical report'' with the slot ``Document'' to ``medical report''.   
\end{itemize}

\begin{algorithm}[ht!]
\caption{Online Inference.} 
\label{alg:Framwork} 
\begin{algorithmic}[1] 
\REQUIRE 
The input utterance $\u$; the mention-concept set $\CS=\bigcup_{r=1}^4 \CS_{r}$, where $r\in\{\Action, \Question, \Argument, \Problem\}$, $\CS_{r}=\{\bigcup_{k_j=1}^{N_k}(m_{k_j}, r, c_{k}), k\in [1, M_r]\}$, $M_r$ is the number of concepts within the intent-role $r$, $N_k$ is the number of mention-concept pairs with concept $c_{k}$; $f$ is a phrase embedding function; $\delta$ is a parameter to filter out dissimilar mentions; $K$ is the number of nearest neighbors; the pattern set $\P$.

\ENSURE The set of intent-role mentions with concept ID, $Result$;

\STATE$\{(m_1, r_1), \cdots, (m_N, r_N)\}\leftarrow \mbox{IRL}(\u)$ 
\STATE {$\{c_1, \ldots, c_N\}\leftarrow \mbox{ConInfer}$($\{(m_1, r_1), \cdots, (m_N, r_N)\}, \CS, \delta, K$)} 
\STATE {$(I, S)\leftarrow \mbox{ISInfer}(\{(m_1, r_1, c_1), \cdots, (m_N, r_N, c_N)\}, \P)$}
\STATE \RET~$(I, S)$
\item[]
\STATE\FUNCTION{~$\mbox{ConInfer}$}{($\{(m_1, r_1), \cdots, (m_N, r_N)\}$,  $\CS$, $\delta$, $K$)} 
\STATE $Result=\{\}$   
\FOR{$j=1$; $j\leq{N}$; $j+\!+$} 
\STATE $Tuple=\{\}$ 
\FOR{all $(m_i,r_{j}, c_{x})$ in $\CS_{r_j}$, $x$ is its concept ID}
\STATE $sim = \cos(f(m_j), f(m_{i}))$ 
\IF {$sim > \delta$}
\STATE $Tuple = Tuple\cup\{(sim, x)\}$
\ENDIF
\ENDFOR
\STATE $Tuple_{o}\leftarrow$ Sort $Tuple$ by $sim$ with decreasing order 
\STATE Get the concept ID list: $ID_{topK} = Tuple_{o}[0:K]$ 
\STATE Find the majority concept ID $q_o$ in set $ID_{topK}$
\STATE $Result = Result\cup\{q_o\}$  
\ENDFOR
\STATE\RET~$Result$
\STATE\ENDFUNCTION
\label{code:fram:ConceptInfer} 
\item[]
\STATE\FUNCTION{~$\mbox{ISInfer}$}{($\{(m_1, r_1, c_1), \cdots, (m_N, r_N, c_N)\}$, $\P$)}
\STATE $Dict=\{\}$, $Slot=\{\}$
\FOR{$j=1$; $j\leq{N}$; $j++$}
\IF {$r_{j}==\Argument$} 
\STATE $SlotD[c_j] = SlotD[c_j] + \{m_j\}$
\ENDIF
\STATE $Dict[r_j]=Dict[r_j] + \{c_j\}$ 
\ENDFOR
\STATE $p = [\mbox{\bf for~} v \mbox{~in~} \P \mbox{\bf~do~if~} \mbox{set}(Dict.keys)==\mbox{set}(v)]$
\STATE $Intent = []$
\FOR{$r$ in $p$}
\STATE $Intent\leftarrow Intent~+~'-'~+~Dict[r]$  
\ENDFOR
\STATE\RET~$(Intent, SlotD)$
\STATE\ENDFUNCTION
\label{code:fram:ISInfer} 
\end{algorithmic} 
\end{algorithm}

\begin{table}[htp]
\caption{Statistics of the curated datasets}
  \label{tab:data_stat}
  \small
  \begin{tabular}{c|l|r |c|c|c}
    \toprule
   \multirow{2}{*}{{\bf Data}} & {\bf Domain } &
   \multicolumn{1}{l|}{\bf No. of} & \multicolumn{1}{l|}{\bf i./m./a.$^a$} & \multicolumn{1}{l|}{\bf |Vocab|/} &
   \multicolumn{1}{l}{\bf No. of} \\
   & {\bf name} & \multicolumn{1}{l|}{\bf utter.$^b$} & \multicolumn{1}{l|}{\bf utter. len.} & \multicolumn{1}{l|}{\bf domain} & \multicolumn{1}{l}{\bf I/S$^c$}\\
    \midrule
   \multirow{9}{*}{\rotatebox[origin=c]{90}{FinD}} & Insurance & 1,008,682 & 5/14/975 & 4,937 & 1,179/53\\
   & Fin. Mgmt. & 851,537 & 5/14/999 & 5,074 & 705/45\\
   & APP Ops. & 311,373 & 5/15/980 & 4,267 & 969/39\\
   & Banking& 263,458 & 5/14/852 & 3,989 & 578/50\\
   & User Info. & 231,982 & 5/13/770 & 3,999 & 419/44\\
   & Health& 122,385 & 5/15/984 & 3,795 & 126/23\\
   & Reward Pts. & 35,556 & 5/15/998 & 3,112 & 279/15\\
   & RE \& Vehicle& 9,741 & 5/13/398 & 2,111 & 109/11\\
   & Others & 133,959 & 5/13/624 & 3,564 & 669/49\\
   \hline
   \multirow{3}{*}{\rotatebox[origin=c]{90}{ECD}} & Commodity & 342,231 & 5/14/168 & 2,189 & 356/28 \\
   & Logistics & 98,720 & 5/13/228 & 1,825 & 229/26 \\
   & Post-sale & 29,616 & 5/17/295 & 1,941 & 116/25 \\
   \hline   
   \multirow{2}{*}
   {\rotatebox[origin=c]{90}{HRD}} 
    & \multirow{2}{*}{Career} & \multirow{2}{*}{64,559} & \multirow{2}{*}{5/14/624} & \multirow{2}{*}{4,206 }& \multirow{2}{*}{341/43 }\\&&&&\\
  \bottomrule
  \multicolumn{5}{l}{$^a$: i./m./a. is equivalent to minimum/average/maximum.}\\
  \multicolumn{5}{l}{$^b$: utter. denotes utterances.}\\
  \multicolumn{5}{l}{$^c$: I/S denotes intents and slots.}
\end{tabular}
\end{table}

\begin{table}[htp]
\caption{Statistics in the test sets}
  \label{tab:testdata_stat}
  {\small
  \begin{tabular}{c|c|c|c|c|c}
    \toprule
   \multirow{2}{*}{{\bf Data}} & 
   \multicolumn{1}{l|}{\bf No. of}&
   \multicolumn{1}{l|}{\bf i./m./a.$^a$} & \multicolumn{1}{l|}{\bf i./m./a.$^a$} & \multirow{2}{*}{\bf |Vocab|} &
   \multicolumn{1}{l}{\bf No. of} \\
   &  \multicolumn{1}{l|}{\bf domains} & \multicolumn{1}{l|}{\bf utter.$^b$} & \multicolumn{1}{l|}{\bf utter. len.} & &
   \multicolumn{1}{l}{\bf I/S$^c$}\\
    \midrule
 FinD & 9 & 2/167/908 & 5/11/25 & 667 & 178/39\\     
 ECD & 3 & 330/500/604 & 5/9/24 & 562 & 90/27 \\      
 HRD & 1 & 1500 & 5/10/27 & 606 & 121/31 \\
  \bottomrule
  \multicolumn{5}{l}{$^a$: i./m./a. is equivalent to minimum/average/maximum.}\\
  \multicolumn{5}{l}{$^b$: utter. denotes utterances.}\\
  \multicolumn{5}{l}{$^c$: I/S denotes intents and slots.}
\end{tabular}
}
\end{table}

\begin{table*}[htp]
 \caption{Compared results on in-domain and out-of-domain SLU datasets.}
   \label{tab:open-world}
   \small
  \begin{tabular}{c|c|c|c|c|c|c}
    \toprule
    \multirow{2}{*}{Dataset} & \multicolumn{2}{c|}{\bf In-domain} & \multicolumn{4}{c}{\bf Out-of-domain} \\
    \cline{2-7} 
    & \multicolumn{2}{c|}{FinD} & \multicolumn{2}{c|}{ECD} & \multicolumn{2}{|c}{HRD} \\ 
    \hline
    Approach & Intent P/R/F1 & Slot P/R/F1  & Intent P/R/F1 & Slot P/R/F1  & Intent P/R/F1 & Slot P/R/F1 \\
    \midrule
    POS & 0.34/0.35/0.34 & 0.49/0.50/0.49 & 0.19/0.20/0.19 & 0.56/0.45/0.50 & 0.05/0.05/0.05 & 0.27/0.40/0.32 \\
    DP & 0.48/0.49/0.48 & 0.51/0.79/0.62 & 0.45/0.46/0.45 & 0.60/0.56/0.58 & 0.28/0.28/0.28 & 0.61/0.41/0.49 \\
    Joint-BERT & 0.89/0.89/0.89 & 0.92/0.90/0.91 & 0.19/0.21/0.20 &  0.49/0.64/0.55 & 0.02/0.02/0.02 &  0.42/0.33/0.37 \\
    RCAP & 0.82/0.84/0.83 & 0.85/0.89/0.87 & 0.79/0.80/0.79 & 0.85/0.79/0.82 & 0.75/0.76/0.75 & 0.84/0.74/0.79 \\
    RCAP + refine & \textbf{0.91/0.90/0.90} & \textbf{0.92/0.91/0.92} & \textbf{0.85/0.86/0.85} & \textbf{0.85/0.81/0.83} & \textbf{0.90/0.91/0.90} & \textbf{0.86/0.75/0.80} \\
  \bottomrule
\end{tabular}
\end{table*}

\section{Experiments}
\label{sec:exp_all}
In this section, we conduct experiments on the AISI task to answer the following questions:
\begin{itemize}
\item[--]  \textbf{Q1:} What is the performance of our RCAP on the AISI task for in-domain and out-of-domain datasets?
\item[--]  \textbf{Q2:} What is the effect of each module in our RCAP and to what extend does our RCAP save human effort in the AISI task?
{\item[--]  \textbf{Q3:} What are the differences between our RCAP and traditional annotation procedure in deriving the intent-slot in real scenarios?}
\end{itemize}

\subsection{Experiment Setup}
\label{sec:data_prep}
\cparagraph{Data} We curate a financial dataset (FinD) with 2.9 million real-world utterances collected from a financial VPA in 9 domains.  We additionally construct a test set of 1,500 annotated utterances by five domain experts with an inter-annotator agreement of the Fleiss’ Kappa larger than 0.75. 

Moreover, to justify the generalization ability of our RCAP, we apply the learned IRL model, concepts, and patterns from FinD to evaluate the model performance on two out-of-domain datasets: a large-scale public Chinese conversation corpus in E-commerce (ECD)~\cite{zhang2018modeling} with the domains of Commodity, Logistics, and Post-sale, and a human resource dataset (HRD) collected from a human resource VPA.  Similarly, we annotate additional 1,500 utterances of each dataset as the test sets.  The detailed statistics of the curated datasets and the test sets are reported in Table~\ref{tab:data_stat} and Table~\ref{tab:testdata_stat}, respectively.

\cparagraph{Compared Methods} We compare the following methods:
\begin{itemize}
\item[--]  {\bf POS}: Ansj\,\footnote{\url{https://github.com/NLPchina/ansj_seg}}, a popular tool for Chinese terminologies segmentation, is first applied to determine the intent-role labels and then derive the corresponding intent-slot following $ConInfer$ and $ISInfer$ in Algo.~1.
\item[--]  {\bf DP}: a dependency parsing toolbox developed by LTP\,\footnote{\url{https://www.ltp-cloud.com}} is applied to determine the intent-role labels and then derive the corresponding intent-slot following $ConInfer$ and $ISInfer$ in Algo.~1. Both labels in POS and DP can not be directly applied for IRL tasks. Therefore, we design some rules to map  POS tags and DP labels to IRL labels (see more details in Appendix~\ref{sec:SI-IRL}).
\item[--]  {\bf Joint-BERT}~\cite{chen2019bert}: a strong supervised model is trained with a Joint-BERT model on another 3.5k annotated utterances by two domain experts to derive the corresponding 48 predefined intents and 43 predefined slots. In the out-of-domain datasets, we include the intent of ``others'' for out-of-schema utterances. It is noted that the number of labeled utterances is sufficient to train good performance; see more results in Appendix~\ref{sec:SI-Joint-BERT}.  
\item[--]  {\bf RCAP}: our proposed RCAP applies a BERT model defined in Eq.~(\ref{eq:irl}) to derive the intent-role labels, CNN embedding with LPA for concept mining, and the Apriori algorithm for pattern mining. 
\item[--]  {\bf RCAP+refine}: the grouped mentions of our RCAP are manually refined by domain experts to pick out outliers and merge them into new concepts based on experts' experience.
\end{itemize}

\cparagraph{Training Details} In training the IRL model, we fine-tune the BERT-base~\footnote{\url{https://github.com/google-research/bert}} model on 6,000 annotated utterances of FinD in 30 epochs by the following settings: a mini-batch size of 32, a maximum sequence length of 128, and a learning rate of $2 \times 10^{-5}$.  For training the CNN embeddings, we leverage TensorFlow implementation of word2vec with CNN on empirically-set common parameters, such as filter windows of 1, 2, 3, 4 with 32 feature maps each, word2vec dimension of 128.  The maximum length of each mention is 16.  The size of the skip window, i.e., the number of sub-words to consider to the left and right of the current sub-words, is 2.  The skip number, i.e., the number of times to reuse an input to generate a labels, is 2.  The sampled number, i.e., the number of negative examples is 64.  For clustering, the number of nearest neighbors in LPA is empirically set to 5. $K$ is 100 for K-means to attain the best performance.  In intent-role pattern mining, the minimum support value is set to 0.05 and the minimum confidence value is 0.1 for the Apriori algorithm. In $ConInfer$, $\delta$ is set to 0.2 and $K=5$.

\cparagraph{Evaluation Metrics} 
In the experiment, {following~\cite{lin2019deep}, we apply Macro-F1 to evaluate the performance of intent-role detection and the standard metrics, precision, recall, and F1, to measure the performance of slot filling.  }
Here, we adopt v-measure, a comprehensive measurement of both homogeneity and completeness~\cite{yin2014dirichlet}, which ranges in 0 to 1, to evaluate the clustering performance.  The larger the v-measure score, the better the clustering result is. 

\subsection{Performance on AISI}
\label{sec:AISI perform}

Table~\ref{tab:open-world} reports the performance of the model trained on FinD and evaluations on all the three test sets to answer \textbf{Q1}.  The results on FinD are to report the in-domain performance while the results on ECD and HRD are to evaluate the out-of-domain performance.  We have the following observations: 
\begin{itemize}
\item[--]  RCAP significantly outperforms the baselines, POS and DP, on all three datasets under the paired $t$-test ($p<0.05$).  The results make sense because POS and DP perform poor in the task of IRL; see more IRL results in Table~\ref{tab:irl-perf}.   

\item[--]  RCAP attains competitive performance to the strong supervised method, Joint-BERT, on FinD.  The results demonstrate the effectiveness of RCAP in handling in-domain SLU.  More significantly, RCAP can easily transfer to other domains and achieves the best performance in ECD and HRD.  On the contrary, Joint-BERT performs poorly, even worse than POS and DP, in ECD and HRD. This is because Joint-BERT does not learn the semantic features in new utterances and may assign the out-of-schema utterances to wrong intents.   
\item[--]  By examining the results on the out-of-domain datasets, RCAP attains at least 76\% higher F1-score on intent detection and 41\% higher F1-score on slot filling than the best baseline.  This implies that RCAP can effectively infer out-of-domain intent-slot. 
\item[--]  RCAP after refinement achieves the best performance, 0.9 Macro-F1 score in intent detection and 0.92 F1-score in slot filling on FinD.  More importantly, the superior performance can be retained in ECD and HRD.  The results show that the refinement can gain 8.4\%-20.0\% improvement on intent detection and 1.3\%-5.7\% improvement on slot filling. 

\end{itemize}

\begin{table}
\setlength{\tabcolsep}{3.5pt}
  \caption{Compared results on IRL performance in FinD shown by Precision/Recall/F1.}
  \label{tab:irl-perf}
  \small
  \begin{tabular}{c|c|c|c}
    \toprule
      & \multirow{1}{*}{\bf $\Action$}   & \multirow{1}{*}{\bf $\Problem$} & \multirow{1}{*}{\bf $\Question$}   \\
    \midrule
    POS & 0.13/0.30/0.18  & 0.02/0.01/0.01 & 0.32/0.13/0.18  \\
    \hline
    DP & 0.58/0.48/0.53  & 0.51/0.56/0.53 & 0.83/0.56/0.67   \\
    \hline
    RCAP & \textbf{0.86/0.87/0.87} & \textbf{0.88/0.83/0.85}  & \textbf{0.94/0.93/0.93}   \\\midrule
    & \multirow{1}{*}{\bf $\Argument$} & \multicolumn{2}{c}{\bf Overall}\\\hline
    POS &0.21/0.24/0.22 & \multicolumn{2}{c}{0.18/0.20/0.19} \\\hline
    DP &  0.74/0.49/0.59 & \multicolumn{2}{c}{0.68/0.51/0.58}\\\hline
    RCAP& \textbf{0.92/0.91/0.92}  & \multicolumn{2}{c}{\textbf{0.91/0.90/0.90}}\\
  \bottomrule
\end{tabular}
\end{table}

\subsection{Drill-down Analysis}\label{sec:drill-down}
In the following, we analyze the effect of each module in RCAP and the schema induction efficiency to answer \textbf{Q2}.

\cparagraph{IRL Performance} {IRL is a crucial step to the success of RCAP.  Here, we evaluate the performance, i.e., precision, recall, and F1-score, on identifying the intent-role labels of utterances and report the performance on  sub-words level. {\bf Overall} shows a weighted score of the corresponding score on all intent-roles.  To provide references, we compare RCAP with the baselines, POS and DP.  Since Joint-BERT cannot provide the intent-role labels, we do not report its result here.
%\label{sec:open_exp}

Table~\ref{tab:irl-perf} reports the performance of compared methods and shows the following observations: 
\begin{itemize} 
\item[--]  RCAP by utilizing BERT significantly outperforms POS and DP in intent-role labeling.  POS and DP perform poorly and attain only 0.19 and 0.58 on overall F1-scores on FinD.  This implies that it is non-trivial to determine the intent-roles. 
\item[--]  By examining the performance on each role, we observe that the F1-scores in $\Action$ and $\Problem$ are slightly lower than those in $\Question$ and $\Argument$.  A reason is that the mentions in both $\Action$ and $\Problem$ contain verbs or verb phrases, which leads to more errors in predicting the roles.  Our trained BERT model can still achieve competitive performance, i.e., 0.87 and 0.85 F1-score, on $\Action$ and $\Problem$, respectively.  
\end{itemize}
}

\begin{figure}%[htb]
  \includegraphics[scale=0.35]{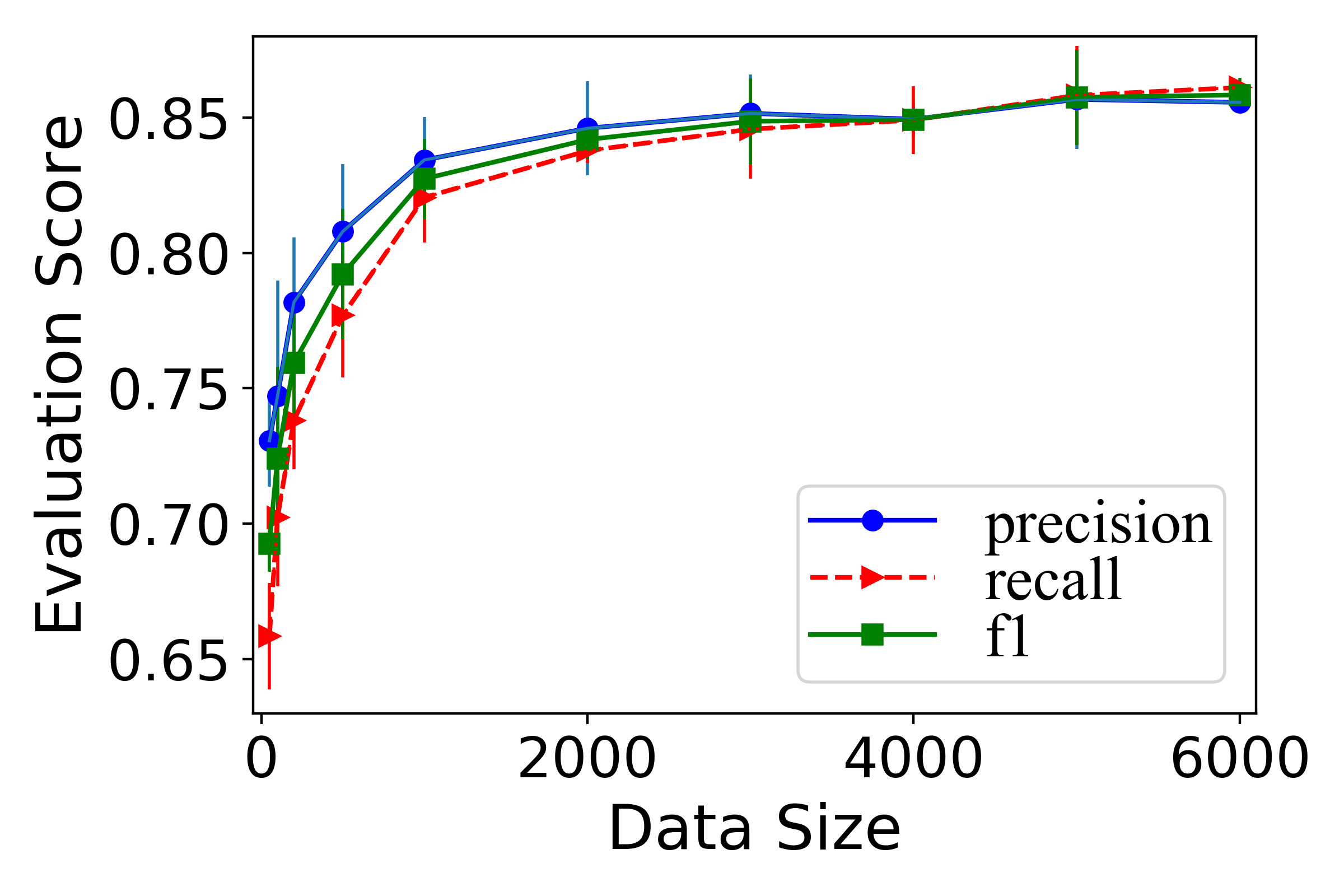}
  \centering
  \caption{Performance of IRL models trained with different number of training data.}
  \label{fig:varying_IRL_training_data}
\end{figure}
\cparagraph{Effect of the number of annotated data in IRL} Here, we argue that to attain satisfactory IRL performance, our RCAP only needs to annotate a small number of utterances.  To support this argument, we ask three domain experts to annotate 6,500 utterances with around 15K mention-role pairs.  In the test, to mimic the real-world scenarios, we keep all utterances without eliminating those with multiple intents or with only $\Argument$.  This makes the test set slightly different from Table~\ref{tab:testdata_stat}.  
In the test, we hold 500 annotated utterances for test while varying the number of training samples in \{50, 100, 200, 500, 1,000, 2,000, 3,000, 4,000, 5,000, 6,000\}.  Figure~\ref{fig:varying_IRL_training_data} shows the total performance of precision, recall, and F1 scores on the sub-word level for all intent roles.  The results show that when the number of training data reaches 2,000, F1 converges to 0.85, which corresponds to 0.9 in Table~\ref{tab:irl-perf}.  The slightly worse performance in Fig.~\ref{fig:varying_IRL_training_data} than that in  Table~\ref{tab:irl-perf} comes from the broader types of utterances in the test set.  Overall, we only need to annotate around 2,000 utterance to attain satisfactory IRL performance, which indicates very light labeling effort.

\begin{figure}[htb]
  \includegraphics[scale=0.35]{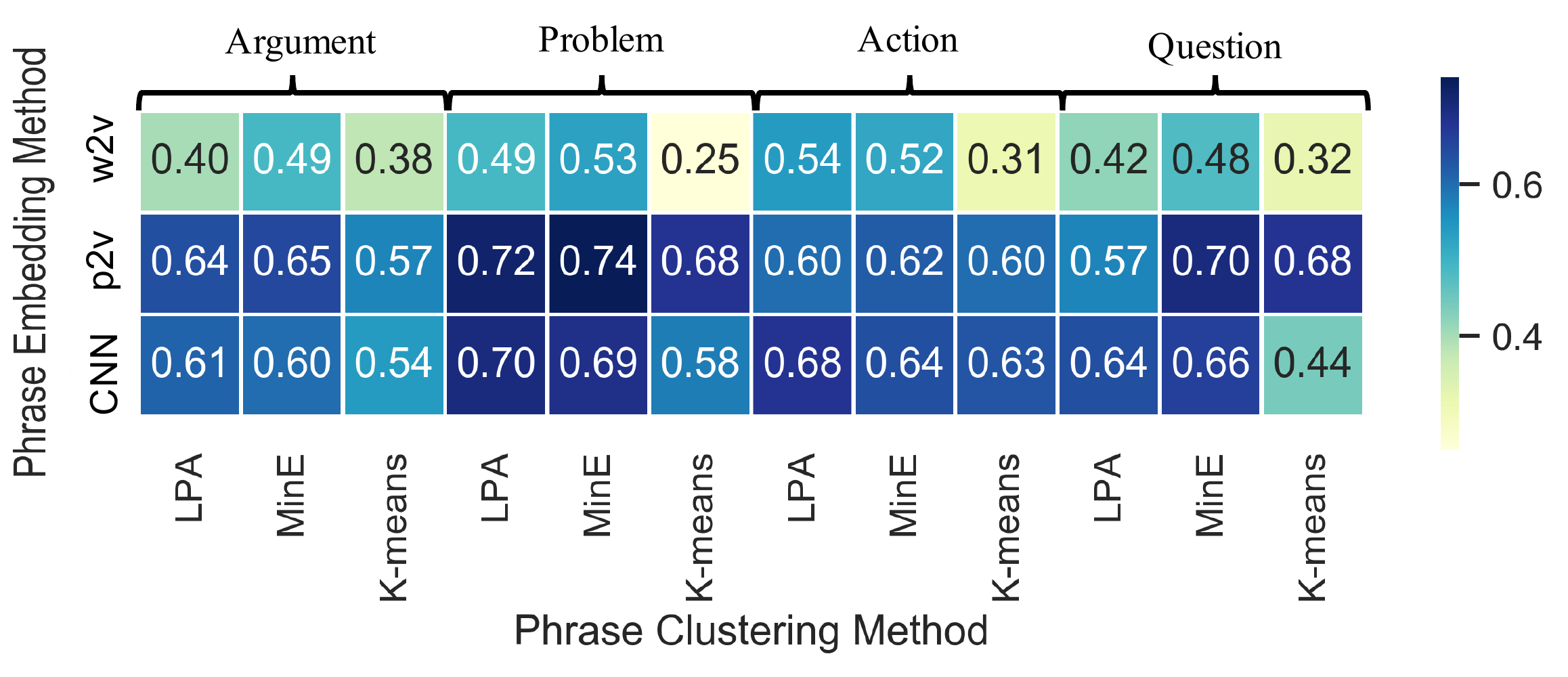}
  \centering
  \caption{Intent-role mention clustering performance.}
  \label{fig:clustering_results}
\end{figure}
\cparagraph{Effect of Concept Mining} \label{sec:exp_concept_clustering}
We test three mention embedding methods and three clustering algorithms as described in Sec.~\ref{sec:concept_mining}, on all 2.9 million utterances in FinD.  We then extract around 10,000  frequent intent-role mentions for evaluations.  The  clustering results in  Fig.~\ref{fig:clustering_results} implies that:  
\begin{itemize}
\item[--]  Among the three mention embedding methods, phrase2vec and CNN significantly outperform words2vec while CNN performs slightly better than phrase2vec.  We conjecture that phrase2vec have effectively absorbed the contextual information for the mentions and CNN embeddings can capture semantic information for the words inside mentions.
\item[--]  Among the three clustering methods, LPA and MinE significantly outperform K-means while LPA is slightly better than MinE. 
We believe that both LPA and MinE capture the structure of the data and contain higher tolerance on embedding errors.  On the contrary, K-means may converge to a local minimum and heavily rely on the initial state. 
\item[--]  The results show that by applying the mention embedding method of phrase2vec or CNN embedding and the clustering algorithm of LPA or MinE, we can yield satisfactory v-measure scores, i.e., in the range of 0.6 to 0.8.  For other metrics, we also obtain similar observations; see more results on Homogeneity and Completeness in Appendix~\ref{sec:SI-clustering}.
\end{itemize}

% It is noted that the above patterns also include intents without $\Argument$.  
\cparagraph{Results of Intent-role Pattern Mining} 
By applying the Apriori algorithm, we can obtain totally 10 patterns written into 5 typical patterns as shown in the results of Pattern Mining in Fig.~\ref{fig:patterns}, $\Action$-$(\Argument)$, $(\Argument)$-$\Question$, $\Action$-$(\Argument)$-$\Question$,  $\Problem$-$(\Argument)$, and $\Problem$-$(\Argument)$-$\mbox{Quest}$-$\mbox{ion}$, where $\Argument$ may not appear in the patterns.  For example, the patterns of $\Action$-$(\Argument)$ and $\Action$-$\Question$-$(\mbox{Argum}$-$\mbox{ent})$ also include the patterns of $\Action$ and $\Action$-$\Question$ without $\Argument$, respectively.  These patterns can cover over 70\% of utterances in FinD.  The remaining utterances consist of multiple intents or chitchats without explicit intents, which are not the targeted cases in this paper. 

\cparagraph{Performance of Concept Inference} 
We test the performance of $ConInfer$ in Algo.~1.  Since frequently-appeared mentions usually can be directly mapped to the corresponding concepts, to test the performance in long-tail situations, we additionally collect 1,000 utterances from FinD.  Each utterance contains at least one low-frequent mention, i.e., appearing less than 20 times in the dataset.  We obtain an inference accuracy of 0.88 on this set, which implies the robustness of our RCAP in concept inference. 

\begin{table}
  \caption{Schema induction performance.}
  \label{tab:schema_induction}
  \small
  \begin{tabular}{cccc}
    \toprule
     & Intents & Slots & Time (hours)\\
    \midrule
    MANUAL & 7 & 16 & \textbf{24} \\
    RCAP  &  16 & 16 & \textbf{2}\\
  \bottomrule
\end{tabular}
\end{table}

\begin{figure}[htb]
  \includegraphics[scale=0.30]{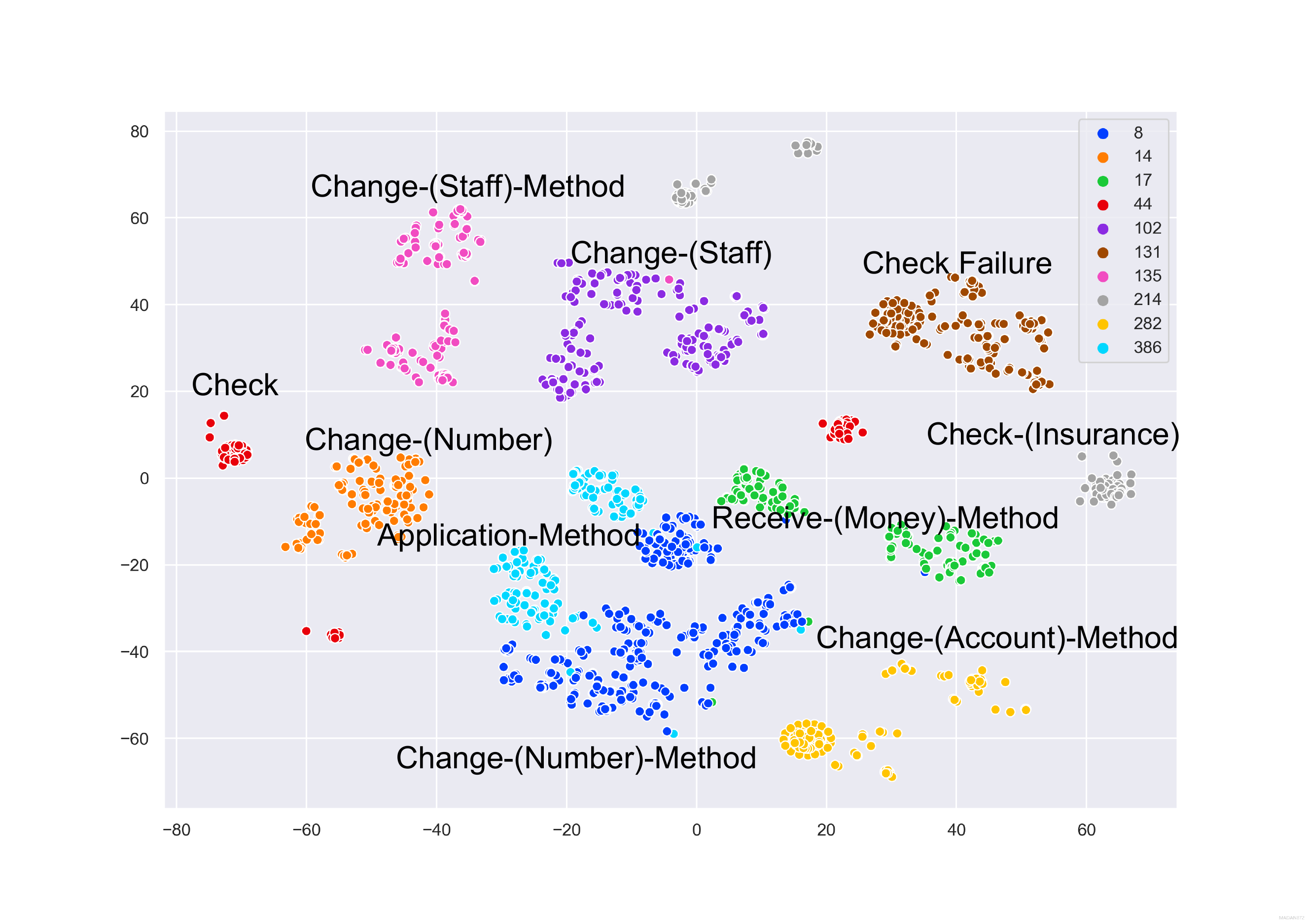}
  \centering
  \caption{Visualizing the clustering arrangement for utterances belonging to ten intents in FinD, where different colors denote different intents.}
  \label{fig:tsne_results}
\end{figure}

\cparagraph{Efficiency} We compare the time cost of traditional manual-schema induction and our RCAP.  We ask a domain expert in ``Health'' domain to derive a schema by the following common schema induction steps:
1) selecting utterances in the ``Health'' domain; 2) grouping similar utterances and abstracting them into intents; 3) enumerating slots for each intent; 4) repeating step 2) and 3) to construct the schema.
In comparison, another expert only needs to determine whether the derived intent-slot labels by our RCAP lie in the ``Health'' domain.   
Since utterances with the same intents may be sparsely distributed in the corpus, the schema induction may be extremely difficult and time consuming.  As shown in Table~\ref{tab:schema_induction}, it takes the first  domain expert about 24 hours to manually derive only 7 intents and 16 corresponding slots.  On the contrary, our RCAP can automatically group utterances of the same intents.  The other expert can, therefore, take only 3 hours to produce a similar schema with 9 additional intents; see more results in Appendix~\ref{sec:SI-schema}.  

\subsection{Case Studies}
We now present the qualitative evaluation of our RCAP to answer \textbf{Q3}.  To show RCAP's ability to conflate different utterances into the same or similar intent category, we concatenate the intent-role mention embeddings of each utterance and plot them by t-SNE in a 2D-space.  Fig.~\ref{fig:tsne_results} shows that the utterances with the same intents (representing by the same colors) cluster in compact groups.  For example, the utterances of ``I want to change my sales manager.'' and ``Change another representative'' are grouped into the intent of ``Change-(Staff)'' in purple, while the utterances of ``How can I have another customer service representative?'' and ``What are the ways to replace my current sales manager'' are grouped into the intent of ``Change-(Staff)-Method'' in pink.  Due to their similar semantic meanings, both intents lie close to each other in Fig.~\ref{fig:tsne_results}.  This implies that our RCAP can effectively group similar utterances into same intents and separate different intents.

Table~\ref{tab:coarse-fine} lists the manual annotation and our RCAP on eight utterances.  Our RCAP can provide more detailed intents comparing with the manual annotated results.  Moreover, our RCAP can induce specific intents for long-tailed utterances.  On the contrary, manual annotation usually assigns them to the intent of \emph{Others}. 

In conclusion, concepts produced by RCAP can help unify utterances with similar semantic meanings into the same intents.  Besides, the detected intents contain fine-grained information and can help induce a meaningful schema, which can cover long-tail utterances.

\begin{table}
  \caption{Fine-grained intents discovered by RCAP vs. manually mined intents for eight cases from FinD.}
  {\small
  \label{tab:coarse-fine}
  \begin{tabular}{@{~}p{3.1cm}|p{2.9cm}|p{1.6cm}}
    \toprule
    \multicolumn{1}{c|}{\bf Utterances} & \multicolumn{1}{|c|}{\bf RCAP} & \multicolumn{1}{|c}{\bf Manual }\\
    \midrule
    {What materials I need to include to loan application} & {(Loan~Application)~$-$~Document~Consultant} 
    & {Loan Application} % \multirow{3}{*}
    \\\cline{1-2}
    {When will the mortgage arrive? } & 
    {Arrival-(Loan,Time)-Consultant} & \\\cline{1-2}
    {My bank loan was rejected.} & {Refused-(Loan)} & \\
    \hline      
   {Can I change the insured of this car insurance?} & {Replace-(Insured)-Feasibility} & % \multirow{2}{*}
   {Insurance Info Modification}\\\cline{1-2}
    {Change the policyholder } & {Replace-(Policyholder)} & \\
   \hline
    {What is the weather today?} & {(Weather,Date)-Inquiry} & {Others}
    \\\cline{1-2}
    {The problem I reported was not solved.} &
    {Not solved-(Issue)} & %\multirow{2}{*}
    \\\cline{1-2}
    {Call the bank customer service. } & {Contact-(Customer~Service)} & \\
  \bottomrule
\end{tabular}
}
\end{table}

\section{Related Work}

\cparagraph{Spoken Language Understanding}
The main goal of an SLU component is to convert the spoken language into structural semantic forms that can be used to generate response in dialogue systems \cite{chen2013unsupervised}.  SLU contains two main sub-tasks: intent classification, which can be treated as a multi-label classification  task~\cite{ravuri2015comparative,kim2017onenet,hashemi2016query}, and slot filling, which can be treated as a sequence labeling task~\cite{mesnil2014using,kurata-etal-2016-leveraging,mesnil2013investigation}.  Recently, joint models for intent detection and slot filling ~\cite{liu2016attention,zhang2016joint,xu2013convolutional,guo2014joint,goo2018slot,qin2019stack,chen2019bert} have received more attention since information in intent labels can further guide the slot filling task. The above approaches require predefined intent-slot schemas and huge labeled data annotated by domain experts to attain good performance.  To alleviate the limits, \citet{vedula2020open} identify domain-independent actions and objects, and construct intents on them.  However, their methods extracted intents that are restricted in action-object form, and cannot fill in slots. This motivates us to explore automatic intent-slot induction in broader scenarios.

\cparagraph{Intent-Slot induction} Traditional SLU systems rely heavily on domain experts to enumerate intent-slot schemas, which may be limited and bias the subsequent data collection~\cite{chen2013unsupervised}.  Hence, many works  propose approaches for automatic intent detection. For example, \citet{xia-etal-2018-zero} propose a zero-shot intent classification model to detect un-seen intents.  Their work is useful for similar domain transfer, but not valid for new domains.  
\citet{lin2019deep,kim2018joint} can only detect if an utterance contains out-of-domain intents. 
Unsupervised  approaches~\cite{tur2011towards,tur2012exploiting,chen2013unsupervised,chen2014leveraging,shi-etal-2018-auto} have been proposed to build models for slot induction and filling.  These papers have applied clustering algorithms to group concepts.  However, their performance still rely on the corresponding domains.  To address this issue, we investigate unsupervised domain-independent methods for both intent and slot. 

\section{Conclusion}
In this paper, we define and explore a new task of automatic intent-slot induction.  We then propose a domain-independent approach, RCAP, which can cover diverse utterances with strong flexibility.  More importantly, our RCAP can be effectively transferred to new domains and sheds light on the development of generalizable dialogue systems.  Extensive experiments on real-word datasets show that our RCAP produces satisfactory SLU results without manually-labeled intent-slot and outperforms strong baselines.  As for the out-of-domain datasets, our RCAP can gain great  improvement than the best baseline.  Besides, RCAP can significantly reduce the human effort in intent-slot schema induction and help to discover new or unseen intent-slot with fine-grained details.  Several promising future directions can be considered: (1) extending the single-intent induction to multi-intents induction; (2) utilizing external well-defined knowledge graph to fine-tune the mined concepts; (3) developing a generalizable dialogue systems based on our RCAP.

\begin{acks}
The authors are grateful to the anonymous reviewers for their insightful feedback, to Xuan Li for helpful discussions during the course of this research. We are also immensely grateful to Bingfeng Luo and Kunfeng Lai for their comments and help on an earlier version of the manuscript.
\end{acks}

\bibliographystyle{ACM-Reference-Format}
\bibliography{www2021}

\clearpage
\appendix
\section{Appendix}

\subsection{Vary Training Data Size for Joint-BERT}
\label{sec:SI-Joint-BERT}
\begin{figure}[htb]
  \includegraphics[scale=0.35]{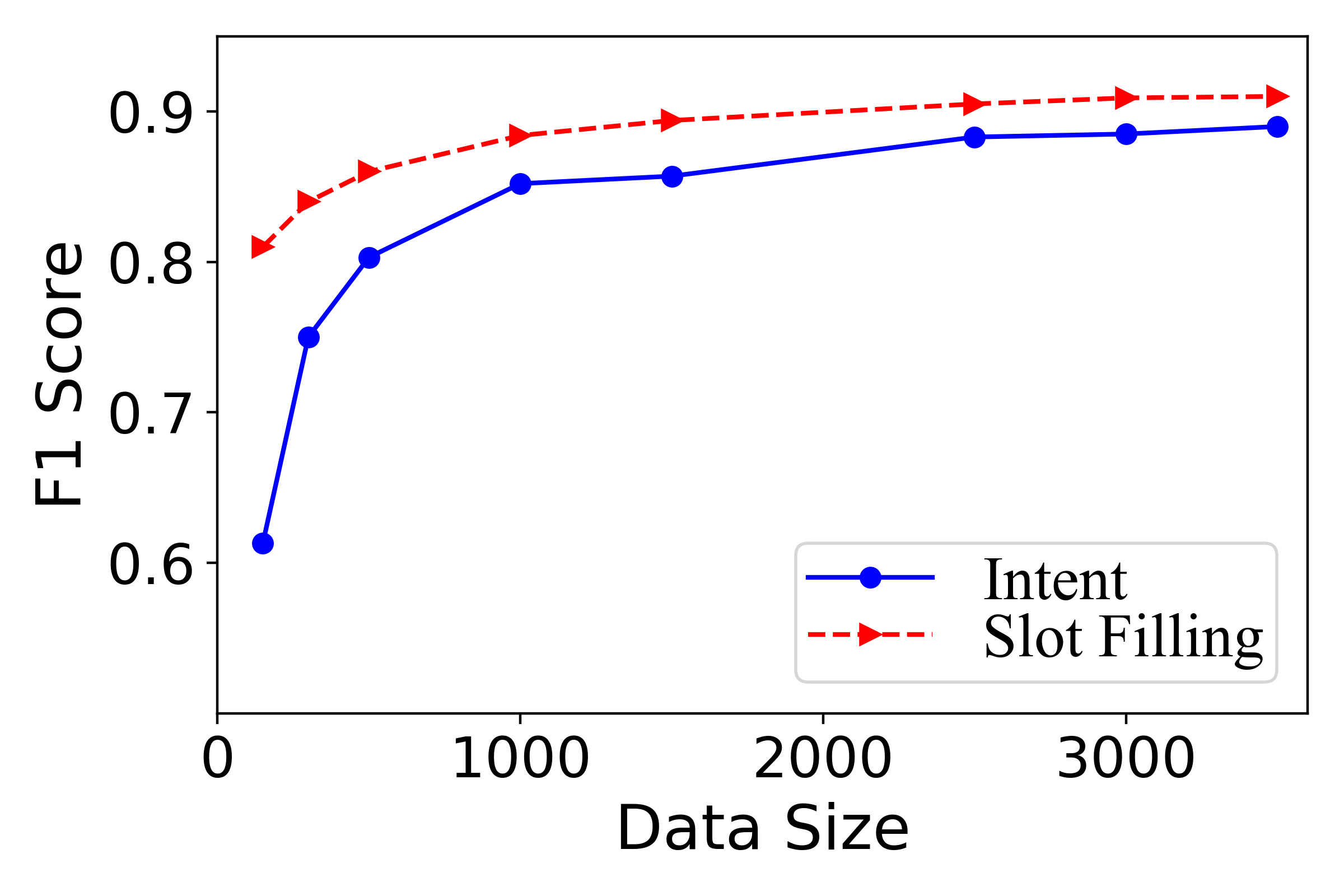}
  \centering
  \caption{Performance of the Joint-BERT model trained on different number of training data, including intent detection (blue) and slot filling (red).}
  \label{fig:DIS-converge}
\end{figure}

In this section, we quantify the annotation requirements for the Joint-BERT model for SLU tasks. We randomly select 5,000 utterances from FinD, and annotate them with intent-slot for test.  
In the test, we hold on additional 1,500 annotated utterances for test while varying the number of training samples in \{150, 300, 500, 1,000, 1,500, 2,500, 3,000, 3,500\}. Figure~\ref{fig:DIS-converge} shows the performance of Joint-BERT in intent detection (blue line) and slot filling (red) on different number of training samples. F1-score is used for evaluations.  The results show that the F1-scores for intent and slot converge to 0.89 and 0.91 when the data size reaches 2,500.   This implies that to train Joint-BERT with good performance, we usually need around 2,500 annotated utterance. 

\begin{figure}[htb]
  \includegraphics[scale=0.30]{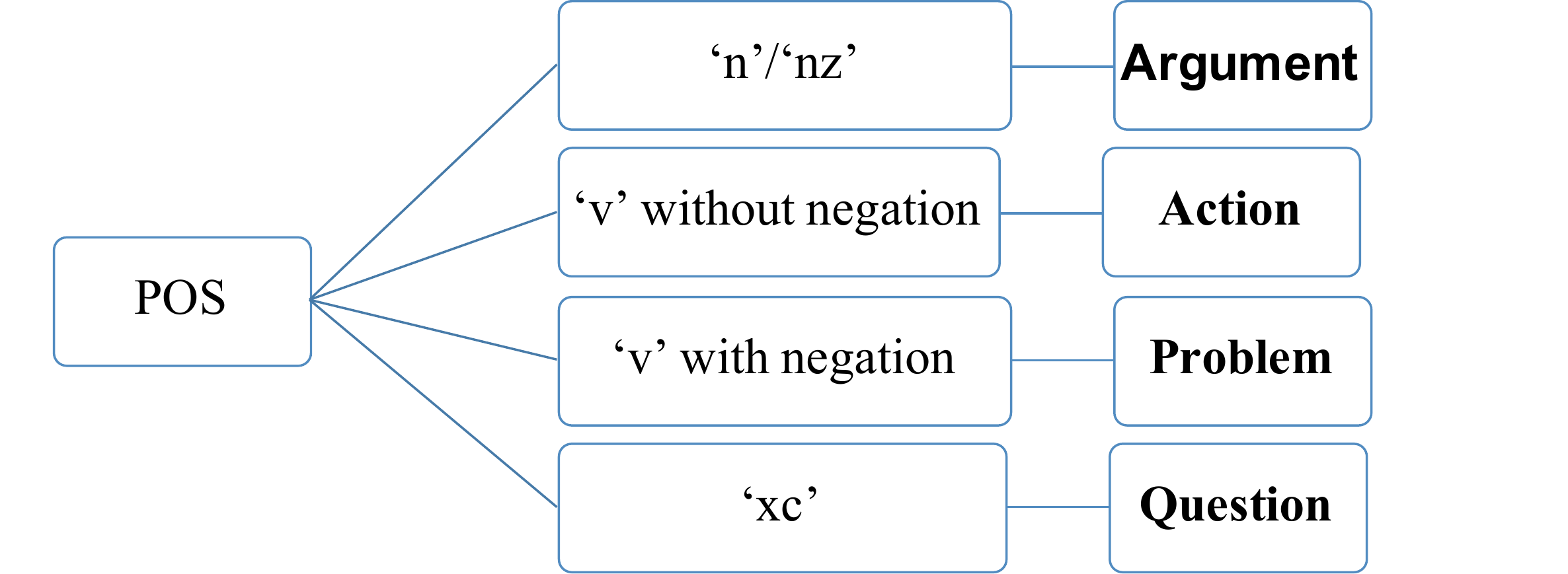}
  \centering
  \caption{Mapping rules from POS tags to IRL intent-role labels.}
  \label{fig:POS-rules}
\end{figure}

\subsection{IRL training Details}
\label{sec:SI-IRL}
We provide more training details for IRL, including POS, DP, and our RCAP.

\begin{itemize}
\item[--]  {\bf POS}: Based on the tags generated by POS, we write a set of rules to derive the corresponding intent-slot labels as in Fig.~\ref{fig:POS-rules}:
\begin{itemize}
\item[--]  A noun or noun phrase is set as $\Argument$;
\item[--]  A verb or verb phrase without negation words is set as $\Action$;
\item[--]  A verb or verb phrase with negation words is set as $\Problem$;
\item[--]  An interrogative word or phrase is set as $\Question$.
\end{itemize}

\item[--]  {\bf DP}: We first use DP on each utterance to generate parsing results with corresponding labels. Afterwards, mention-concept sets mined by RCAP are utilized to decide the intent-role for each mention. 
\begin{itemize} 
\item[--]  lies in $\Action$ mentions and labeled as ``HED" or ``COO" is set as $\Action$;
\item[--]  lies in $\Problem$ mentions and labeled as ``HED" or ``COO" is set as $\Problem$;
\item[--]  lies in $\Argument$ mentions and has dependency relations to $\Action$ or $\Problem$ is set as $\Argument$;
\item[--]  lies in $\Question$ mentions and has dependency relations to $\Action$ or $\Problem$ is set as $\Question$.
\end{itemize}

\end{itemize}

\begin{figure}
  \includegraphics[scale=0.35]{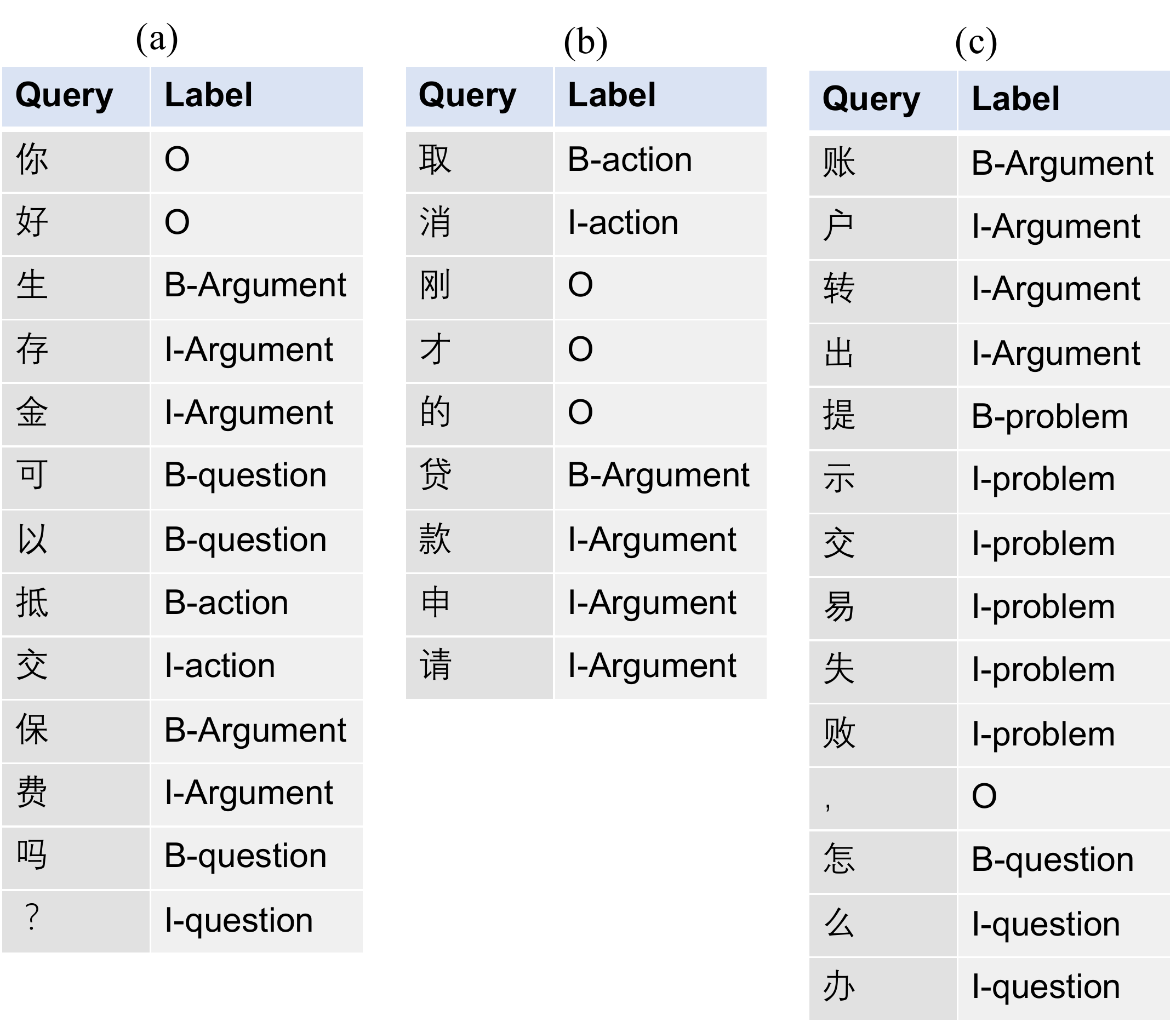}
  \centering
  \caption{Three IRL results: (a) Hi, can the survival benefit pay the premium? (b) Cancel the loan application (c) Account transfer transaction failed, what should I do. The Query columns are sub-words of input utterances to the IRL labeling model, the Label columns are outputs by RCAP. ``B'' indicates the beginning tag, ``I'' indicates the inside tag, and ``O'' indicates the outside tag. }
  \label{fig:SI-0}
\end{figure}

\subsection{Clustering Performance}
\label{sec:SI-clustering}
\begin{figure}
  \includegraphics[scale=0.35]{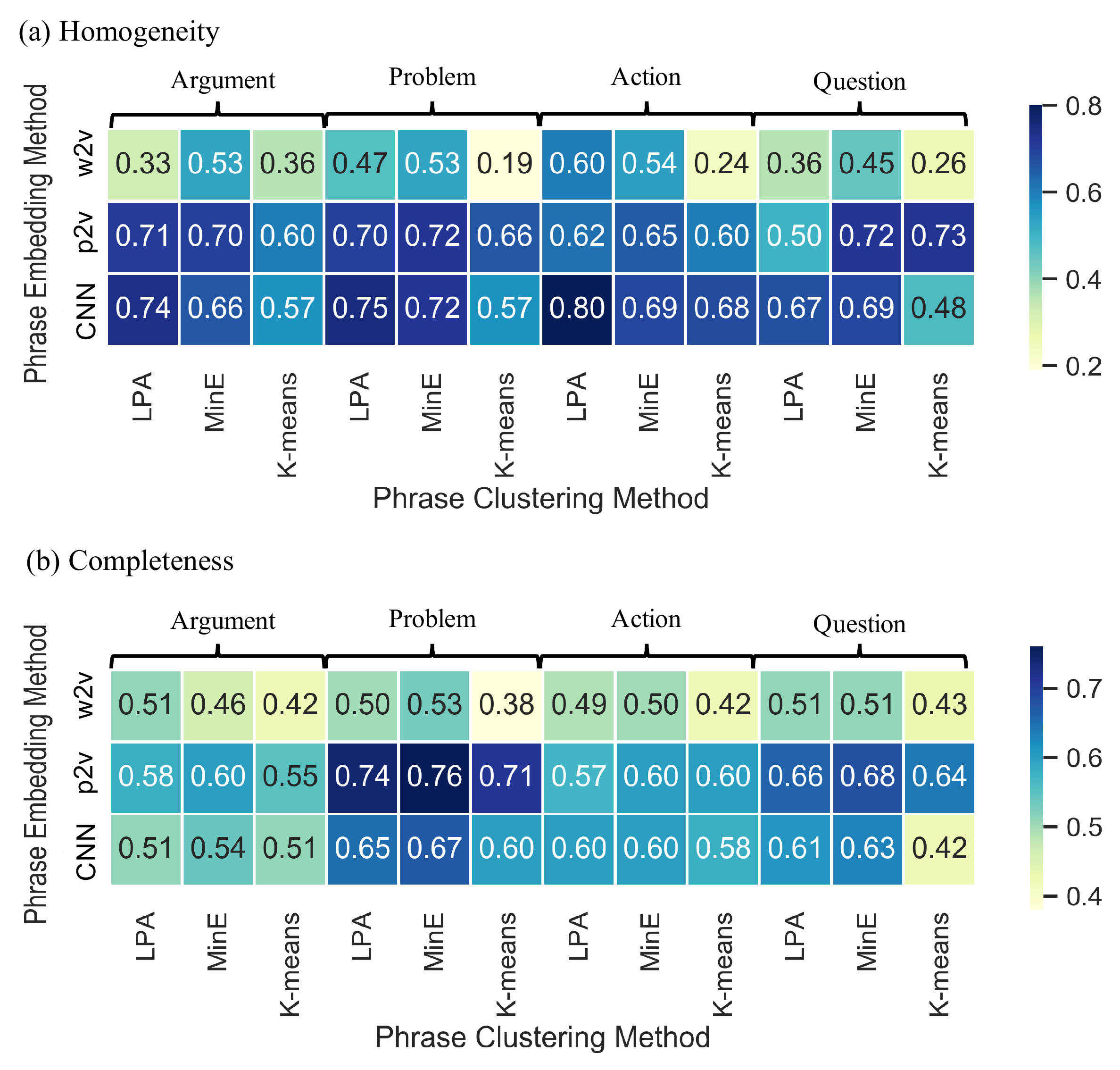}
  \centering
  \caption{
Intent-role mention clustering evaluations on all intent-roles using (a) completeness and (b) homogeneity, with different mention embedding methods and clustering algorithms included. The color bar shows evaluation scores decrease from dark blue to light yellow.}
  \label{fig:clustering}
\end{figure}

We introduce the other two metrics to evaluate the performance of intent-role mention clustering in experiments: homogeneity and completeness.
Homogeneity and completeness are entropy-based measures that are symmetrical to each other. A clustering result satisfies prefect homogeneity if the class distribution with each cluster is skewed to a single class. A clustering result satisfies perfect completeness if all datapoints that are members of a single class are clustered to a single cluster \cite{rosenberg2007v}. V-measure used in  Sec.~\ref{sec:drill-down} is a comprehensive measurement of both homogeneity and completeness \cite{yin2014dirichlet}.  Fig.~\ref{fig:clustering} shows the clustering performance on all intent-roles using homogeneity and completeness. %  as in 

\subsection{Samples in Concept Repository}
\label{sec:SI-concept-repo}
\begin{figure}
  \includegraphics[scale=0.31]{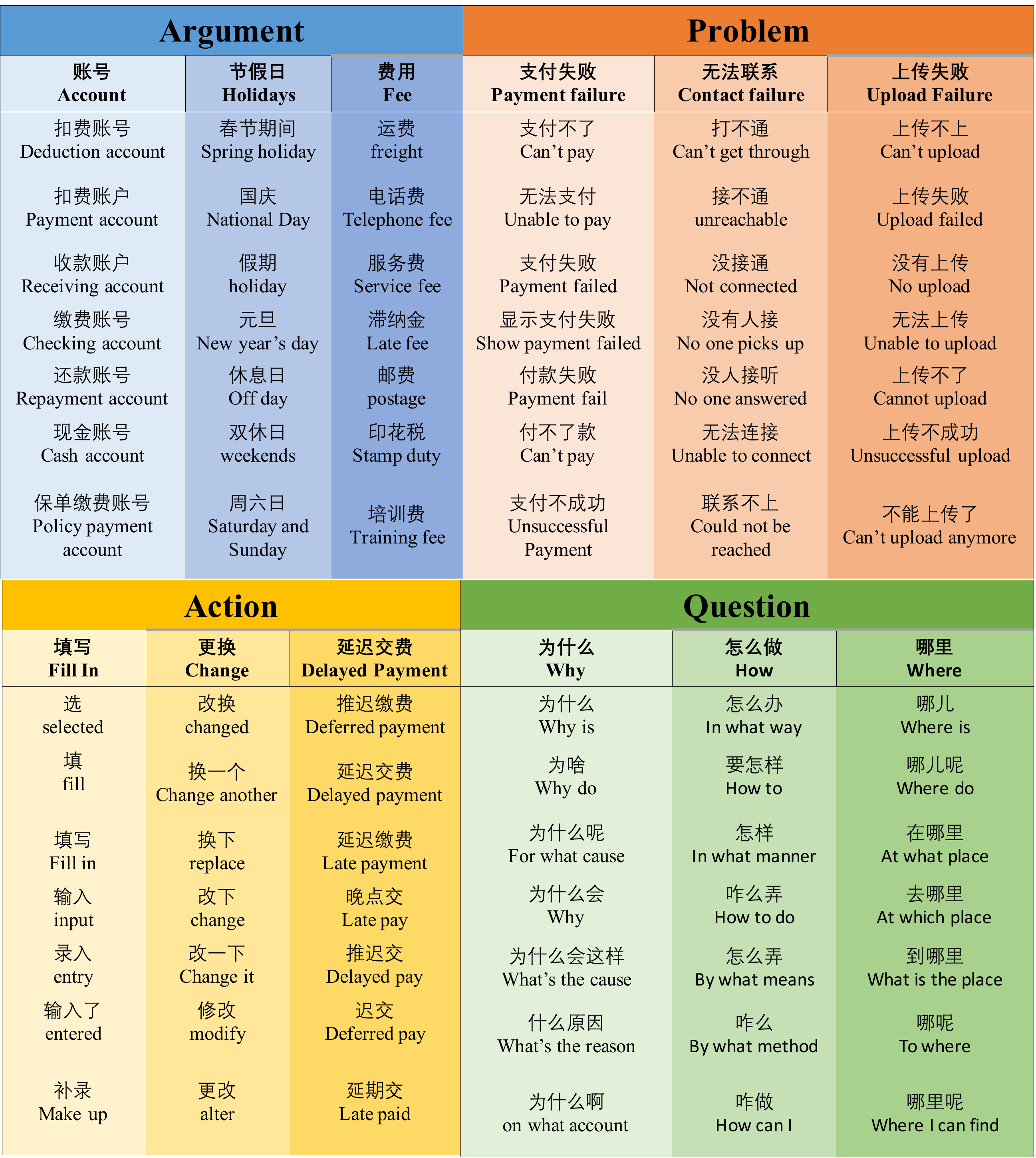}
  \centering
  \caption{Typical concepts for all four intent-roles: $\Argument$, $\Problem$ , $\Action$, and $\Question$. 
  }
  \label{fig:concept-repo}
\end{figure}

In elucidating the concept repository produced by concept mining, we show some  samples for all different intent-roles as in Fig.~\ref{fig:concept-repo}.

\subsection{Schema Induction Results}
\label{sec:SI-schema}
\begin{figure}
  \includegraphics[scale=0.25]{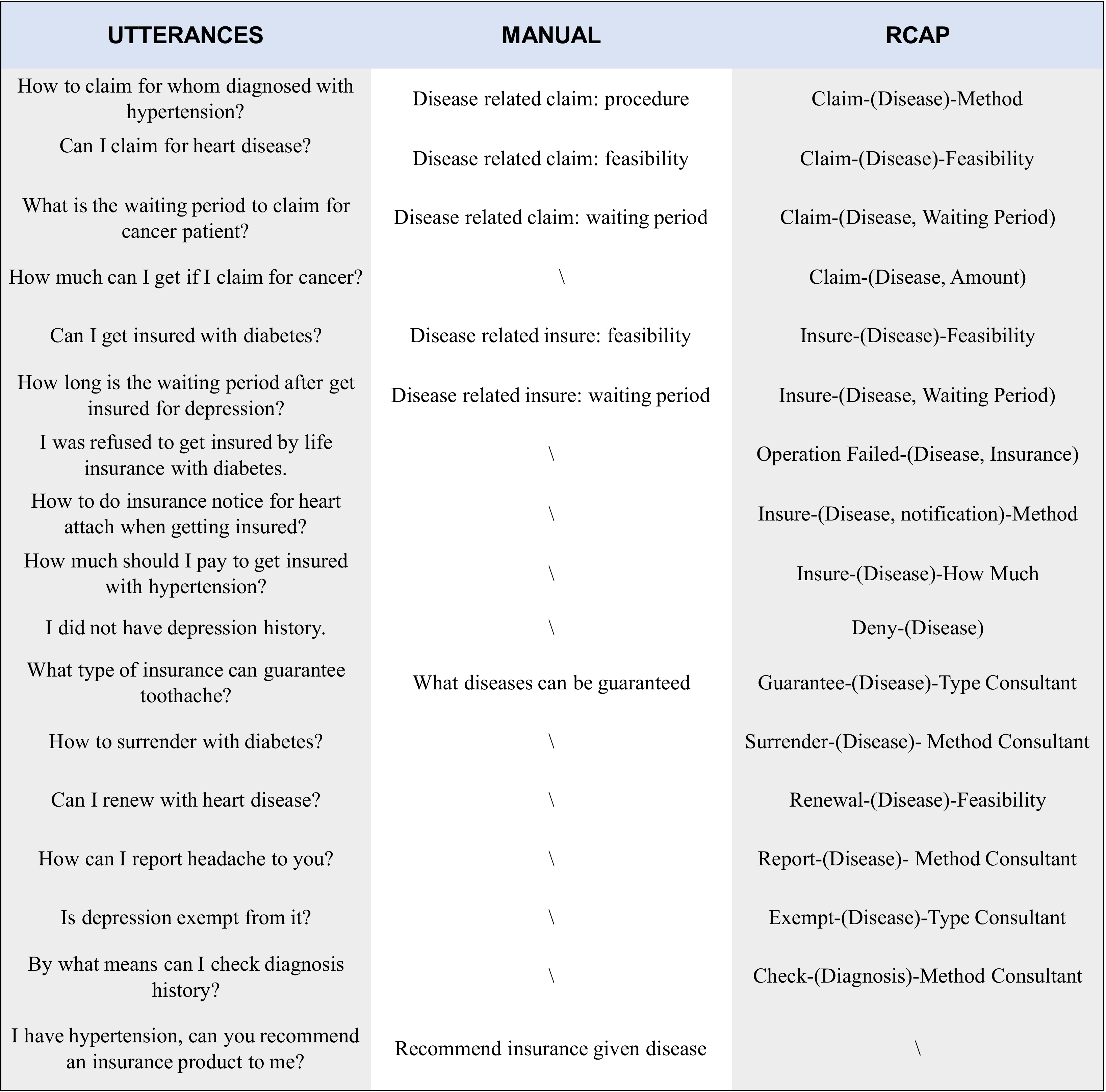}
  \centering
  \caption{
  Results of manual induction and automatic induction by RCAP on the ``Health'' domain.
  }
  \label{fig:schema}
\end{figure}

We enumerate the induced intents in Sec.~\ref{sec:drill-down} by the manual procedure and our RCAP and show them in Fig.~\ref{fig:schema}.

\subsection{Concepts Expansion}
\label{sec:SI-concept-exp}
Though concepts are mined from a large-scale open-domain corpus, they may be still missed in new domains.  To enrich the concepts and  enhance the feasibility of RCAP induced schemas in new domains, we will conduct concept expansion on new utterances when they come from new domains.  

Here, we first extract the corresponding intent-role mentions from the utterances.  After that, we assign these mentions to existing concepts by directly matching.  If they are not matched, we conduct concept inference on them.  {Hence, there exist some mentions that cannot be inferred to a certain concept with high confidence since they may not have enough qualified neighbors; see details in $ConInfer$ function of Algo.~1. Here, we tune $\delta$ and $K$ to control the confidence level for concepts expansion. We then collect these uncategorized mentions and feed them to \emph{concept mining} process to find new concepts. Due to the simple structure of RCAP, new concepts are incremental to the original concept schema.}  

\subsection{Dataset Release}
We release a sample set with 324 utterances from ECD.
%\footnote{ \url{https://www.dropbox.com/s/l90odrlk8og2smw/ecd_open_sample_324.xlsx?dl=0}}.
All other datasets are under review and the desensitized datasets will be released after publication.

\end{document}